\documentclass[runningheads]{llncs}

\makeatletter
\usepackage[bookmarks,unicode,colorlinks=true]{hyperref}%
   \def\@citecolor{blue}%
   \def\@urlcolor{blue}%
   \def\@linkcolor{blue}%

\def\orcidID#1{\smash{\href{http://orcid.org/#1}{\protect\raisebox{-1.25pt}{\protect\includegraphics{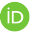}}}}}
\makeatother

\usepackage{amssymb,amsmath,amsfonts}
\usepackage{temporal}
\usepackage{graphicx}
\usepackage{fontawesome} % for fancy envelope symbol
\usepackage{pifont}
\usepackage{prism}
\usepackage{listings}
\usepackage{textcomp}
\usepackage[mathscr]{eucal}
\usepackage{hyperref}
\usepackage{amssymb}
\usepackage{enumitem}
\usepackage{svg}
\usepackage{tikz}

\usetikzlibrary{automata,positioning,decorations.markings,arrows,intersections,shapes,%
  shapes.symbols,calc,backgrounds,shadings}
\usetikzlibrary{overlay-beamer-styles}

\tikzset{
  >=stealth,
  box state/.style={draw,rectangle,minimum size=8mm},
  prob state/.style={draw,very thick,shape=circle,darkblue,minimum size=3mm,inner sep=0mm},
  node distance=2cm,on grid,auto, initial text=,
  every loop/.style={shorten >=0pt},
  accepting/.style={double distance=1.2pt, outer sep = 0.6pt+\pgflinewidth},
  accepting dot/.style={above=-2.5pt,circle,fill,darkgreen,inner sep=2pt,radius=1pt},
  loop above/.append style={every loop/.append style={out=120, in=60, looseness=6}},
  loop below/.append style={every loop/.append style={out=300, in=240, looseness=6}},
  loop left/.append style={every loop/.append style={out=210, in=150, looseness=6}},
  loop right/.append style={every loop/.append style={out=30, in=330, looseness=6}},
  accepting arc/.style={dashed},
  marked/.style={
    dashed,
    opacity=0.3
  },
  marked on/.style={alt=#1{marked}{}},
}

\lstdefinestyle{compact}{basicstyle=\scriptsize,xleftmargin=2ex,numbersep=6pt}
\lstdefinestyle{semicompact}{basicstyle=\footnotesize\ttfamily,xleftmargin=2ex,numbersep=6pt}
\lstdefinestyle{inline}{basicstyle=\sffamily}

\definecolor{darkgreen}{rgb}{0,0.6,0}
\definecolor{lightblue}{rgb}{0.5,0.6,1.0}
\definecolor{lightgray}{rgb}{0.98,0.98,0.98}
\definecolor{mauve}{rgb}{0.58,0,0.82}
\definecolor{sienna}{rgb}{0.6,0.18,0.09}
\colorlet{darkblue}{blue!60!black}
\colorlet{darkred}{red!50!black}
\colorlet{safecellcolor}{yellow!5}
\colorlet{goodcellcolor}{green!10}
\colorlet{badcellcolor}{blue!10}

\newcommand{\M}{Mungojerrie}

\begin{document}
\title{\M: Reinforcement Learning of Linear-Time
  Objectives\thanks{This work was supported in part by the Engineering and Physical Sciences Research Council through grant EP/P020909/1 and by the National Science Foundation through grant 2009022.}}
\author{Ernst Moritz Hahn\inst{1}\orcidID{0000-0002-9348-7684}
  \and Mateo Perez\inst{2}\orcidID{0000-0003-4220-3212}
  \and Sven Schewe\inst{3}\orcidID{0000-0002-9093-9518}
  \and Fabio Somenzi\inst{2}\orcidID{0000-0002-2085-2003}
  \and Ashutosh Trivedi\inst{2}\orcidID{0000-0001-9346-0126}
  \and Dominik Wojtczak\inst{3}\orcidID{0000-0001-5560-0546}}

\institute{
  University of Twente, The Netherlands
  \and
  University of Colorado Boulder, USA
  \and
  University of Liverpool, UK
}
\authorrunning{E. M. Hahn, M. Perez, S. Schewe, F. Somenzi, A. Trivedi, and D. Wojtczak}
% \authorrunning{\M: Reinforcement Learning of Linear-Time Objectives}
%
\maketitle              % typeset the header of the contribution
\begin{abstract}
Reinforcement learning synthesizes controllers without prior knowledge of the system. At each timestep, a reward is given. The controllers optimize the discounted sum of these rewards.

Applying this class of algorithms requires designing a reward scheme, which is typically done manually. The designer must ensure that their
intent is accurately captured. This may not be trivial, and is prone to error. An alternative to this manual programming, akin to programming directly in assembly, is to specify the objective in a formal language and have it ``compiled'' to a reward scheme.

\M{} (\href{https://plv.colorado.edu/mungojerrie/}{plv.colorado.edu/mungojerrie}) is a tool for testing reward schemes for $\omega$-regular objectives on finite models. The tool contains reinforcement learning algorithms and a probabilistic model checker. \M{} supports models specified in PRISM and $\omega$-automata specified in HOA.

\end{abstract}
\section{Introduction}
\label{sec:intro}
Reinforcement learning (RL \cite{Sutton18}) has seen a surge of impressive results in recent years \cite{Silver16,openai2019rubiks,Mnih15}. 
RL agents explore a potentially unknown environment, receiving rewards which provide feedback on performance. The agent then seeks to maximize performance. This process is known as ``learning.'' 
For environments which are unknown, RL is a particularly attractive technique due to its ability to optimize without needing to explicitly construct a model internally.
Applying RL requires converting the objective of the problem into one that can be optimized via RL; a reward function must be designed. 

Currently, the most common way to design the reward function is by hand. This is prone to error \cite{rlblogpost,Wiewio10}. Instead, one may consider writing the objective in a formal language and have it converted into a reward function. A natural choice for this language is Linear Temporal Logic (LTL) \cite{Manna91,Pnueli81}, or more generally, $\omega$-regular languages \cite{Perrin04}. 

$\omega$-regular languages describe infinite sequences. If an infinite run of the system is a word in the $\omega$-regular language, then the property is said to have been satisfied. A valid reward scheme for $\omega$-regular objectives must be such that the optimal strategies for reinforcement learning are guaranteed to be optimal strategies for the $\omega$-regular objective. 

\M{} provides the capability to test this by providing the tools to learn reinforcement learning strategies with performance statistics, and test their optimality with respect to the $\omega$-regular objective with a model checker. The tool supports finite state and action models specified in PRISM \cite{Kwiatk11}, with $\omega$-automata in HOA \cite{Babiak15}.

\begin{figure}[t]
  \begin{center}
    \begin{tikzpicture}
      \begin{scope}[xshift=4cm,yshift=3cm,scale=1,
        action/.style={->,very thick,darkred},
        cell name/.style={circle,fill=white,inner sep=1pt}]
        \coordinate (t1) at (0,0);
        \coordinate (t2) at ($ (t1) + (2,0) $);
        \coordinate (t3) at ($ (t1) + (60:2) $);
        \coordinate (c0) at ($ (t1) + (2,{2/sqrt(3)}) $);
        \foreach \x in {1, 2, 3} {
          \pgfmathparse{\x == 1 ? "badcellcolor" : "goodcellcolor"}
          \edef\cellcolor{\pgfmathresult}
          \path[fill=\cellcolor] (t\x) -- ++(2,0) -- ++(120:2) --cycle;
          \coordinate (c\x) at ($ (t\x) + (1,{1/sqrt(3)}) $);
        }
        % Cell contours.
        \draw[thick] (0,0) -- ++(4,0) -- ++(120:4) --cycle;
        \draw[fill=safecellcolor] (2,0) -- ++(60:2) -- ++(-2,0) --cycle;
        % Cell names.
        \node[cell name] at ($ (t1) + (2,3) $) {$3$};
        \node[cell name] at ($ (t2) + (-0.6,1.5) $) {$0$};
        \node[cell name] at ($ (t1) + (0.4,0.25) $) {$1$};
        \node[cell name] at ($ (t2) + (1,1.3) $) {$2$};
        % Actions.
        \draw[action] ($ (c0) + (0.1,0) $) -- node[pos=0.25,right] {$a$} +(0,0.9);
        \draw[action] ($ (c0) - (0,0.2) $) -- node[pos=0.4,right]
             {$b$} ++(0,-0.5) -- node[at end,below,black] {$1-p$} ++(-0.7,0);
             \draw[action] ($ (c0) - (0,0.2) $) ++(0,-0.5) --
             node[at end,below,black] {$p$} ++(0.7,0);
             \draw[action] ($ (c3) - (0.1,0) $) -- node[pos=0.25,left] {$c$} +(0,-0.9);
             \draw[action] (c1) -- node[pos=0.1,above] {$d$} ++(30:0.9);
             \draw[action] ($ (c2) + (-30:0.2) $)
             arc[radius=0.4,start angle=150,end angle=490] node[at start,right] {$e$};
             \draw[action] (c2) -- node[pos=0.1,above] {$f$} ++(150:0.9);
      \end{scope}
      \begin{scope}[yshift=4.5cm,scale=0.9,every
        node/.style={scale=0.9}, transform shape]
        % Nodes.
        \node[state,initial,fill=safecellcolor] (G) {$\mathtt{safe}$};
        \node[state,fill=badcellcolor] (B) [right=2cm of G]
        {$\mathtt{trap}$};
        % Transitions.
        \path[->]
        (G) edge [loop above] node[accepting dot,label={$\mathtt{g} \wedge \neg \mathtt{b}$}] {} ()
        edge [loop below] node {$\neg\mathtt{g} \wedge \neg \mathtt{b}$} ()
        edge              node {$\mathtt{b}$} (B)
        (B) edge [loop below] node {$\top$} ();

      \end{scope}
      \node[inner sep=0pt] (MJ) at (0.6,-0.1)
      {\includegraphics[trim=0cm 20cm 0cm 0cm, clip, height=2cm]{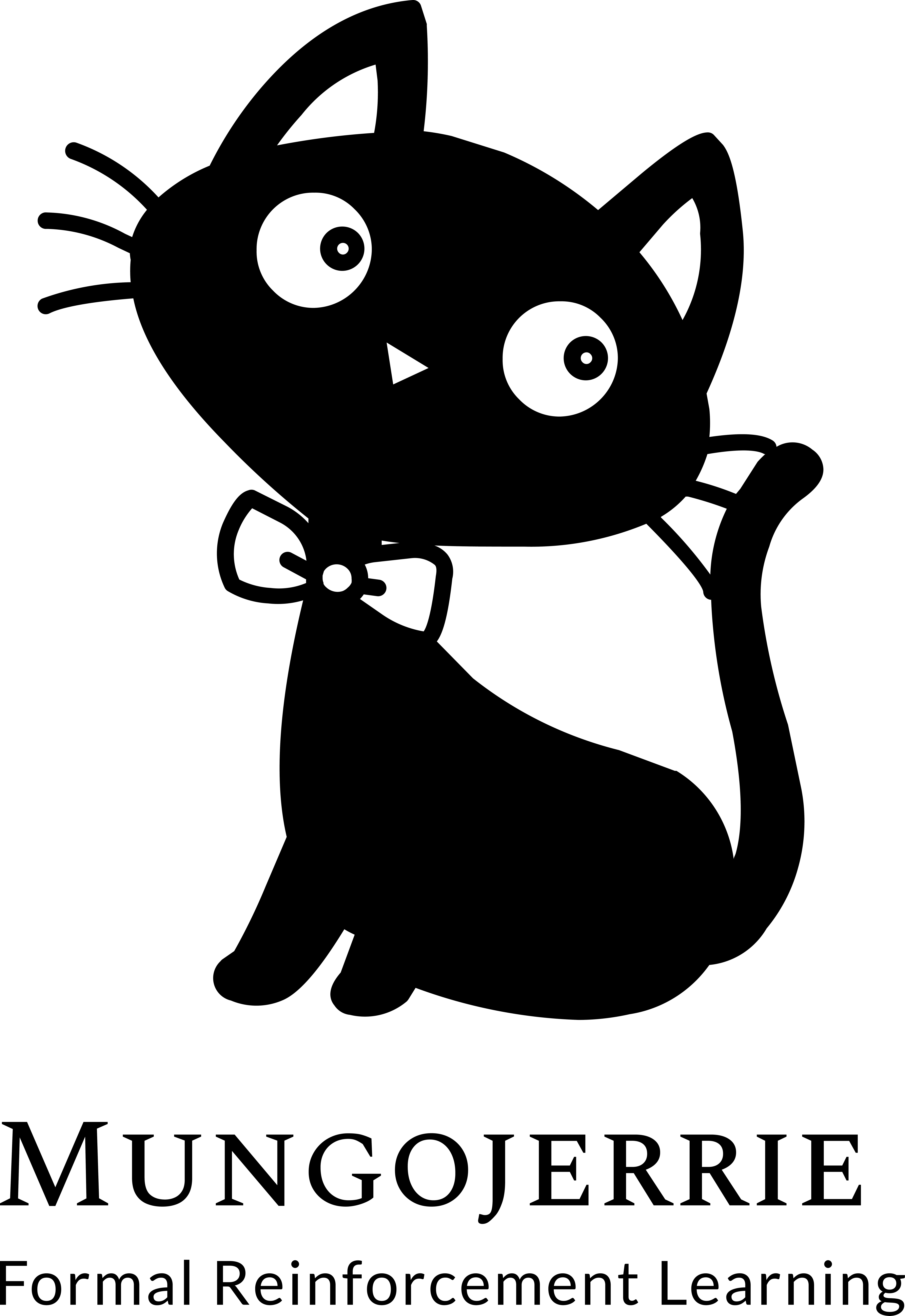}};
      \node (Interpreter) [left=1.5cm of MJ] {Interpreter};
      \node[single arrow, fill=blue!25] at (3.15,0.1) {Observation, Reward};
      \node[single arrow, rotate=40, shape border
      rotate=180,fill=blue!25] at (2.7,1.9) {\phantom{longe}State\phantom{longe}};
      \node[single arrow, fill=blue!25, rotate=90] at (6,1.75) {Action};
      \node[double arrow, fill=blue!25, rotate=90] at (0.7,1.9)
      {\phantom{state}};
      \node[inner sep=0pt] (Robot) at (6,0)
      {\includegraphics[height=2cm]{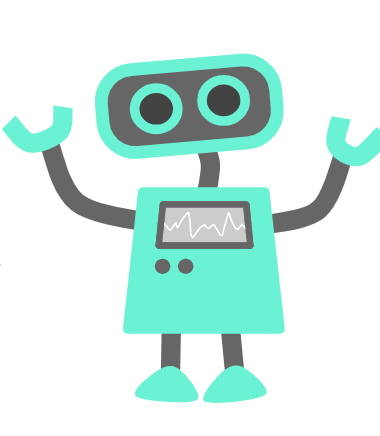}};
      \node (Learner) [right=1.5cm of Robot] {Agent};
      \node (Objective) at (1.8,5.5) {Objective};
      \node (MDP) at (7.5,5.5) {Model};
    \end{tikzpicture}
  \end{center}
  \caption{The reinforcement learning loop implemented within Mungojerrie. The interpreter assigns reward to the agent based on the state of the model and automaton.}
  \label{fi:overview}
\end{figure}
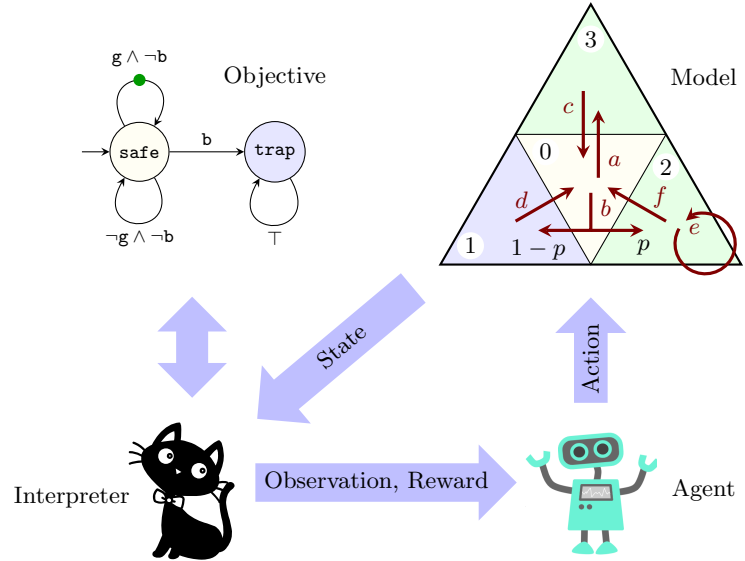

\begin{figure}
  \centering
\begin{lstlisting}[language=prism,firstline=4,numbers=left]
mdp

const int Wealth = 5;   // initial gambler's wealth
const double p   = 1/2; // probability of winning one bet

label "rich" = x = 7;
label "poor" = x = 1;

module gambler
  x : [0..7] init Wealth;

  [b0] x=0 | x=7 -> true; // absorbing states
  [b1] x>0 & x<7 -> p : (x'=x+1) + (1-p) : (x'=x-1);
  [b2] x>1 & x<6 -> p : (x'=x+2) + (1-p) : (x'=x-2);
  [b3] x>2 & x<5 -> p : (x'=x+3) + (1-p) : (x'=x-3);
endmodule
\end{lstlisting}
  \caption{A Gambler's Ruin model in the PRISM language.  Line 13, for
  example, says that when $1 < x < 6$, the gambler may bet two units
  because action \texttt{b2} is enabled.  The `$+$' sign does double
  duty: as addition symbol in arithmetic expressions and as separator
  of probabilistic transitions.}
  \label{fi:gambler}
  \vspace{5mm}
  \begin{tikzpicture}
    % Nodes.
    \node[state,initial above,fill=safecellcolor] (Q0) {$0$};
    \node[state,fill=goodcellcolor] (Q1) [right=3.5cm of Q0] {$1$};
    \node[state,fill=badcellcolor]  (Q3) [below=2cm of Q1] {$3$};
    \node[state,fill=safecellcolor] (Q2) [below=2cm of Q0] {$2$};
    % Transitions.
    \path[->]
    (Q0) edge [loop left] node {$\neg(\mathtt{rich} \vee \mathtt{poor})$} ()
    edge node [accepting dot,label={$\mathtt{rich}$}] {} (Q1)
    edge [swap] node {$\neg\mathtt{rich} \wedge \mathtt{poor}$} (Q2)
    (Q1) edge [loop right] node[accepting dot,label={$\top$}] {} ()
    (Q3) edge [loop right] node {$\top$} ()
    (Q2) edge [loop left] node {$\neg(\mathtt{rich} \vee \mathtt{poor})$} ()
    edge node [accepting dot,label={right: $\mathtt{rich}$}] {} (Q1)
    edge [swap] node {$\neg\mathtt{rich} \wedge \mathtt{poor}$} (Q3);
  \end{tikzpicture}
  \caption{Deterministic B\"uchi automaton equivalent to the LTL formula
    $\neg\mathtt{poor} \until \bigl(\mathtt{rich} \vee
    (\mathtt{poor} \wedge \nextt (\neg\mathtt{poor} \until
    \mathtt{rich}))\bigr)$.  The transitions marked with the green
    dots are accepting.}
  \label{fi:gambler-dba}
  \vspace{5mm}
  \includegraphics[width=\textwidth,clip=true,trim=6cm 0cm 1cm 0cm]{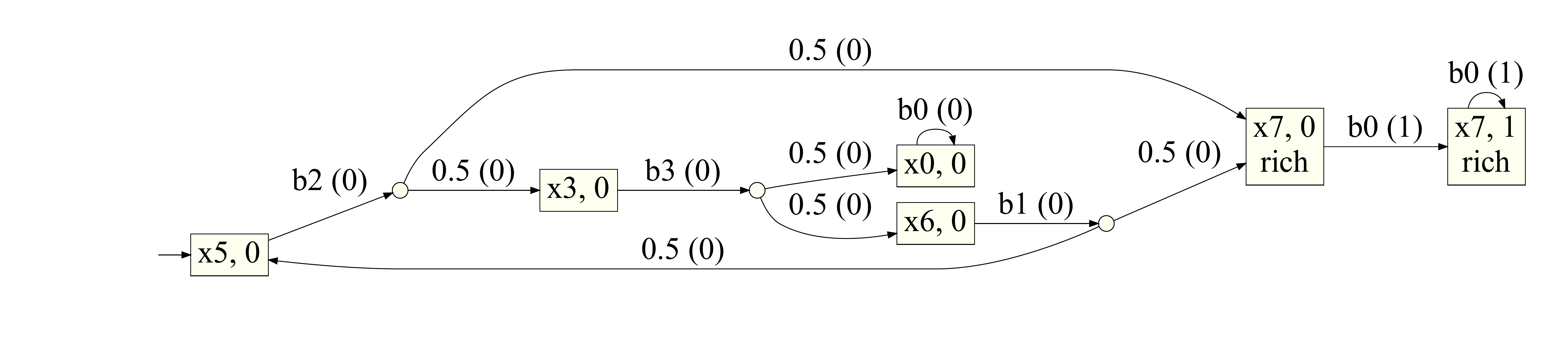}
  \caption{Optimal gambler strategy for the objective of
    Fig.~\protect{\ref{fi:gambler-dba}}.  Boxes are decision states
    and circles are probabilistic choice states.  For a decision
    state, the label gives the value of $x$ and the state of the
    automaton.  Transitions are labelled with either an action or a
    probability, and with the priority ($1$ for accepting and $0$ for
    non-accepting).}
  \label{fi:gambler-strategy}
\end{figure}

Figure~\ref{fi:gambler} shows an example Markov Decision Process in
which a gambler places bets with the aim of accumulating a wealth of
$7$ units.  In addition the gambler will quit if her wealth wanes to
just one unit more than once.  This objective is captured by the
(deterministic) B\"uchi automaton of Fig.~\ref{fi:gambler-dba}.  

\M{} computes a strategy for the gambler that maximizes the probability of
satisfaction of the objective.
Figure~\ref{fi:gambler-strategy} shows the Markov chain that results
from following this strategy. This figure was minimally modified from GraphViz \cite{EllsonGKNW04} output from \M{}.  Note that the strategy altogether
avoids the state in which $x=1$; hence it achieves the same
probability of success ($5/7$) as an optimal strategy for the simpler
objective of eventually reaching $x=7$ (without going broke).
\M{} computes the strategy of Fig.~\ref{fi:gambler-strategy} by
reinforcement learning; it can also verify it by probabilistic model
checking.

\section{Overview of \M{}}
\label{sec:overview}

\subsection{Models}
The model for the systems used in \M{} consist of finite sets of states and actions, where states are labelled with atomic propositions. There are at most two strategic players: Max player and Min player. Each state is controlled by one player. We call models where all states are controlled by Max player Markov Decision Processes (MDPs) \cite{Put94}. Else, we refer to them as stochastic games \cite{Condon92}.

\M{} supports parsing models specified in the PRISM language. The allowed model types are ``mdp'' (Markov Decision Process) and ``smg'' (Stochastic Multiplayer Game) with two players.  There should be one initial state. The interface for building the model is exposed, allowing extensions of \M{} to connect with parsers for other languages.

\subsection{Properties}
Our properties are $\omega$-regular languages. Starting from the initial state, the players produce an infinite sequence of states with a corresponding infinite sequence of atomic propositions, an $\omega$-word. The inclusion of this $\omega$-word in our $\omega$-regular language determines whether this particular run satisfies the property or not. Max player is trying to maximize the probability that a run is satisfying while Min player is minimizing.

We specify our $\omega$-regular language as an $\omega$-automaton, which may be nondeterministic. For model checking and reinforcement learning, this nondeterminism must be resolved on the fly. Automata where this can be done in any MDP without changing acceptance are said to be Good-for-MDPs (GFM) \cite{Hahn20}. Automata where this can be done in any stochastic game without changing acceptance are said to be Good-for-games (GFG) \cite{Henzin06}.

In general, nondeterministic B\"uchi automata are not GFM, but two classes of GFM B\"uchi automata with limited nondeterminism have been studied: suitable limit-deterministic B\"uchi automata \cite{Hahn15,Sicker16b} and slim B\"uchi automata \cite{Hahn20}.

The user of \M{} can either provide the $\omega$-automaton directly or use one of the supported external translators to generate the automaton from LTL with a single call to \M{}. \M{} reads automata specified in the HOA format. The supported LTL translators are \textsc{ePMC} plugin (see Section~\ref{sec:epmc}), \textsc{Spot} \cite{duret.16.atva2}, and Owl \cite{Kretin18} for generating slim B\"uchi, deterministic parity, and suitable limit-deterministic B\"uchi automata. The user is responsible for the $\omega$-automata provided directly having the appropriate property, GFM or GFG.

For use in \M{}, the automata must have labels and acceptance
conditions on the transitions (not on the states). The acceptance
conditions supported by \M{} should be reducible to parity acceptance
conditions without altering the transition structure of the
automaton. This includes parity, B\"uchi, co-B\"uchi, Streett 1 (one
pair), and Rabin 1 (one pair) conditions.  Nondeterministic automata
must have B\"uchi acceptance conditions.  Generalized acceptance
conditions are not supported in version 1.0.

\subsection{Reinforcement Learning}
In reinforcement learning, we have a model as described before with
the addition of a reward function. The reward function
probabilistically assigns a reward $R_{t+1} \in \mathbb{R}$ dependent
on the state and action at timestep $t$. As our players move through
the model, we produce a sequence of states, actions, and rewards which
we index for each timestep $(S_0, A_0, R_1, S_1, A_1, R_2,
\ldots)$. The objective in reinforcement learning is to solve 
$$\max_\pi \min_\nu \mathbb{E}_{\pi,\nu} \left[ \sum_{t=0}^\infty
  \gamma^t R_{t+1} \right] \enspace,$$
where $\pi$ is the strategy for Max player, $\nu$ is the strategy for Min player, $\gamma \in [0,1)$ is the discount factor, and $R_t$ is the reward at timestep $t$. We can set $\gamma = 1$ when with probability $1$ we enter an absorbing sink (termination) where we receive no reward. This is called the episodic setting.

Version 1.0 of \M{} includes the stochastic game extensions of
Q-learning \cite{WD92}, Double Q-learning \cite{vanHas10}, and Sarsa($\lambda$) \cite{Sutton88} for reinforcement learning in finite state and action models. We collectively refer to parameters which are set by hand prior to running a reinforcement learning algorithm as hyperparameters. \M{} supports changing all hyperparameters from the command line. As the design of \M{} separates the learning agent(s) from the reward scheme, extending \M{} to include another reinforcement learning algorithm is easy.

\subsection{Reward Schemes}

The user of \M{} can either select one of
the reward schemes included with the tool or extend the tool to
include a new reward scheme.  The following reward schemes are included in version 1.0 of \M{}:
\begin{itemize}
  \item Reward from the PRISM specification.
  \item The reward scheme from \cite{Hahn19}.
  \item The reward scheme from \cite{DBLP:journals/corr/abs-1909-07299}.
  \item The reward schemes from \cite{Hahn20c}.
  \item The reward schemes from \cite{Hahn20b}.
  \item The reward scheme from \cite{Sadigh14}. 
\end{itemize}
Note that the reward scheme of \cite{Sadigh14} may produce sub-optimal
strategies \cite{Hahn19}.  All other schemes guarantee that a strategy
that maximizes the expected return of the reward also maximizes the
probability that the $\omega$-regular objective be satisfied.
The separation of the learning agent(s) from the reward scheme makes
the inclusion of new schemes easy. The primary effort here will be to
modify the construction of the model passed to the model checker if
there are additional states beyond those due to the original model and automaton.
\section{Tool Design}
\label{sec:design}

\begin{figure}[t]
    \begin{center}
    \begin{tikzpicture}
    \tikzset{inputoutput/.style={trapezium, draw, inner sep=3pt, trapezium
        right angle=-60pt, trapezium left angle=60pt, align=center}}
    \tikzset{data/.style={draw, inner sep=3pt, align=center}}
    \tikzset{code/.style={draw, rectangle, inner sep=3pt, align=center}}
  
    \node[inputoutput] (ltl) at (-1, 4) {LTL};
    \node[inputoutput] (hoa) [below=2.5 of ltl] {HOA};
    \node[inputoutput] (prism) [right=6 of hoa] {PRISM};
    \node[code] (spot) [below=1.25 of ltl] {SPOT};
    \node[code] (epmc-mj) [left=2 of spot] {\textsc{ePMC} Plugin};
    \node[code] (owl) [right=1.25 of spot] {Owl};
    \node[code] (ext) [right=1.25 of owl, minimum height=0.43cm] {$\dots$};

    \node[code] (hparser) [below=1.5 of hoa] {HOA Parser};
    \node[code] (pparser) [below=1.5 of prism] {PRISM Parser};
    \node[inputoutput] (parity) [below=1 of hparser] {Automaton};
    \node[inputoutput] (model) [below=1 of pparser] {Model};
    \node[code] (const) [right=3 of parity] {Product\\Construction};
    \node[inputoutput] (prod) [below=1.25 of const] {Product};

    \node[code] (mc) [below=1.25 of prod] {Model Checking \\ + \\ Game Solver};
    \node[code] (gym) [below right=0.5 and 3 of prod] {Interpreter \\ (Gym)};
    \node[code] (learner) [below=1.5 of gym] {Agent(s) \\ (Learner)};

    \node[inputoutput] (strat) [below=2.5 of mc] {Strategy + Values};
    \node[inputoutput] (qtable) [right=3 of strat] {Q-table(s)};

    \draw[->] (ltl) -- ($(ltl)+(0,-0.625)$) -- ($(ltl)+(-2,-0.625)$) -- (epmc-mj);
    \draw[->] (ltl) -- ($(ltl)+(0,-0.625)$) -- ($(ltl)+(1.25,-0.625)$) -- (owl);
    \draw[->] (ltl) -- ($(ltl)+(0,-0.625)$) -- ($(ltl)+(2.5,-0.625)$) -- (ext);
    \draw[->] (ltl) -- (spot);
    \draw[->] (epmc-mj) -- ($(epmc-mj)+(0,-0.625)$) -- ($(epmc-mj)+(2,-0.625)$) -- (hoa);
    \draw[->] (owl) -- ($(owl)+(0,-0.625)$) -- ($(owl)+(-1.25,-0.625)$) -- (hoa);
    \draw[->] (ext) -- ($(ext)+(0,-0.625)$) -- ($(ext)+(-2.5,-0.625)$) -- (hoa);
    \draw[->] (spot) -- (hoa);

    \draw[->] (hoa) -- (hparser);
    \draw[->] (prism) -- (pparser);
    
    \draw[->] (hparser) -- (parity);
    \draw[->] (pparser) -- (model);
    \draw[->] (parity) -- (const);
    \draw[->] (model) -- (const);
    \draw[->] (const) -- (prod);
    \draw[->] (prod) -- (mc);
    \draw[->] (parity) -- ($(parity)+(0,-0.625)$) -- ($(parity)+(6,-0.625)$) -- (gym);
    \draw[->] (model) -- (gym);
    \draw[<->] (gym) -- (learner);
    \draw[<->] ($(mc.east)$) -- ($(mc.east)+(0.4,0)$) -- ($(gym.west)+(-0.46,0)$) -- ($(gym.west)$);

    \draw[<->] (learner) -- (qtable);
    \draw[->] ($(learner)+(-0.25,-0.46)$) -- ($(learner)+(-0.25,-0.75)$) -- ($(learner)+(-2.75,-0.75)$) -- ($(strat)+(0.25,0.25)$);
    \draw[->] (mc) -- (strat);

    \draw[dashed] (-4.5, 1) rectangle (2.5, 4.5);
    \draw (-3,0.5) rectangle (7,-5.25);
    
    \end{tikzpicture}
    \end{center}
    \caption{\label{fig:flowchart}\M{}'s block diagram}
\end{figure}

\M{} begins its execution by parsing the input PRISM and HOA. (See
upper part of Fig.~\ref{fig:flowchart}.)  The HOA
is either read in from a file or piped from a call to one of the
supported LTL translators. In particular an LTL translator capable of
producing slim B\"uchi automata \cite{Hahn20} comes with the tool. See
Section~\ref{sec:epmc} for details. Requested automaton modifications,
such as determinization, are run after this step.

If specified, \M{} creates the synchronous product between the
automaton and the model, and runs model checking or game solving. The requested strategy and values are returned. Algorithms used at this step can be found in \cite{deAlfa98,Hahn16,Hahn17}.

If learning has been specified, the interpreter takes the automaton and model, without explicitly forming the product, and provides an interface akin to \cite{Brockm16} for the reinforcement learning agent to interact with the environment and receive rewards. When learning is complete, the Q-table(s) can be saved to a file for later use with \M{}, and the interpreter forms the Markov chain induced by the learned strategy and sends it the model checker for verification.

\M{} is written in C++. It has been tested on Ubuntu 20.04, Ubuntu 18.04, and MacOS 11.2. 

\subsection{Slim B\"uchi Automata Generation}
\label{sec:epmc}

\newcounter{nodeno}
\newcommand{\nnode}[1]{\refstepcounter{nodeno}\label{#1}}
\newcommand{\nodename}[1]{\nnode{#1}(\ref{#1})}
\newcommand{\refnode}[1]{(\ref{#1})}

\begin{figure}
  \begin{center}
  \begin{tikzpicture}
  \tikzset{inputoutput/.style={trapezium, draw, inner sep=3pt, trapezium
      right angle=-60pt, trapezium left angle=60pt, align=center}}
  \tikzset{data/.style={draw, internaldata, inner sep=3pt, align=center}}
  \tikzset{code/.style={draw, rectangle, inner sep=3pt, align=center}}

  \node[inputoutput] (ltl) at (-1.5,0.5) {LTL formula \nodename{ltl}};
  \node[inputoutput] (hoai) at ( 1.5,0.5) {HOA file \nodename{hoai}};
  \node[code] (ltl2ntba) at (-1.5,-1) {translate (\textsc{Spot}) \nodename{ltl2ntba}};
  \node[code] (parsehoa) at (1.5,-1) {parse \nodename{parsehoa}};
  
  \node[inputoutput] (genba) at (0,-2) {NTLBA \nodename{genba}};
  \node[code] at (-3,-3) (genslim) {construct SBA \nodename{genslim}};
  \node[code] at ( 3,-3) (genld) {construct LDBA \nodename{genld}};
  \node[inputoutput] at (-3,-6) (sba) {SBA \nodename{sba}};
  \node[inputoutput] at ( 3,-4) (ldba) {LDBA \nodename{ldba}};
  \node[code] at ( 3,-5) (minld) {minimize LDBA \nodename{minld}};
  \node[inputoutput] at ( 3,-6) (mindld) {minimized LDBA \nodename{mindld}};
  \node[code] at ( 3,-7) (consgame) {construct simulation game \nodename{consgame}};
  \node[inputoutput] at ( 3,-8) (game) {simulation game \nodename{game}};
  \node[code] at (3,-9) (solver) {game solver \nodename{solver}};
  \node[inputoutput] at ( 0,-11) (hoao) {HOA file \nodename{hoao}};

  \node at (5,-1.25) {$\textsc{ePMC}$ plugin};
  
  \draw[->] (ltl) -- (ltl2ntba);
  \draw[->] (ltl2ntba) |- (genba);
  \draw[->] (hoai) -- (parsehoa);
  \draw[->] (parsehoa) |- (genba);
  \draw[->] ($(genba.south)+(-0.5,0)$) |- (genslim);
  \draw[->] ($(genba.south)+(-0.25,0)$) -- ($(genba.south)+(-0.25,-8.45)$);
  \draw[->] ($(genba.south)+( 0.25,0)$) |- (genld);
  \draw[->] (genslim) -- (sba);
  \draw[->] (genld) -- (ldba);
  \draw[->] (ldba) -- (minld);
  \draw[->] (minld) -- (mindld);
  \draw[->] (genba) |- (consgame);
  \draw[->] (mindld) -- (consgame);
  \draw[->] (consgame) -- (game);
  \draw[->] (game) -- (solver);
  \draw[->] (mindld) -- ($(mindld)+(3,0)$) -- ($(mindld)+(3,-3.75)$) -- ($(mindld)+(-3,-3.75)$) -- ($(mindld)+(-3,-4.74)$);
  \draw[->] ($(sba.south)$) -- ($(sba.south)+(0,-3.5)$) -- ($(sba.south)+(2.5,-3.5)$) -- ($(sba.south)+(2.5,-4.45)$);
  \draw[dashed] ($(solver.west)+(0,0)$) -- node[above] {won} ($(solver.west)+(-1.95,0)$);
  \draw[dashed] ($(solver.east)+(0,0)$) -- node[above] {lost} ($(solver.east)+(1.7,0)$);

  \draw (-5,-0.25) rectangle (7,-10.25);
  
  \end{tikzpicture}
  \end{center}
  \caption{\label{fi:epmc}Automata generation block diagram}
\end{figure}
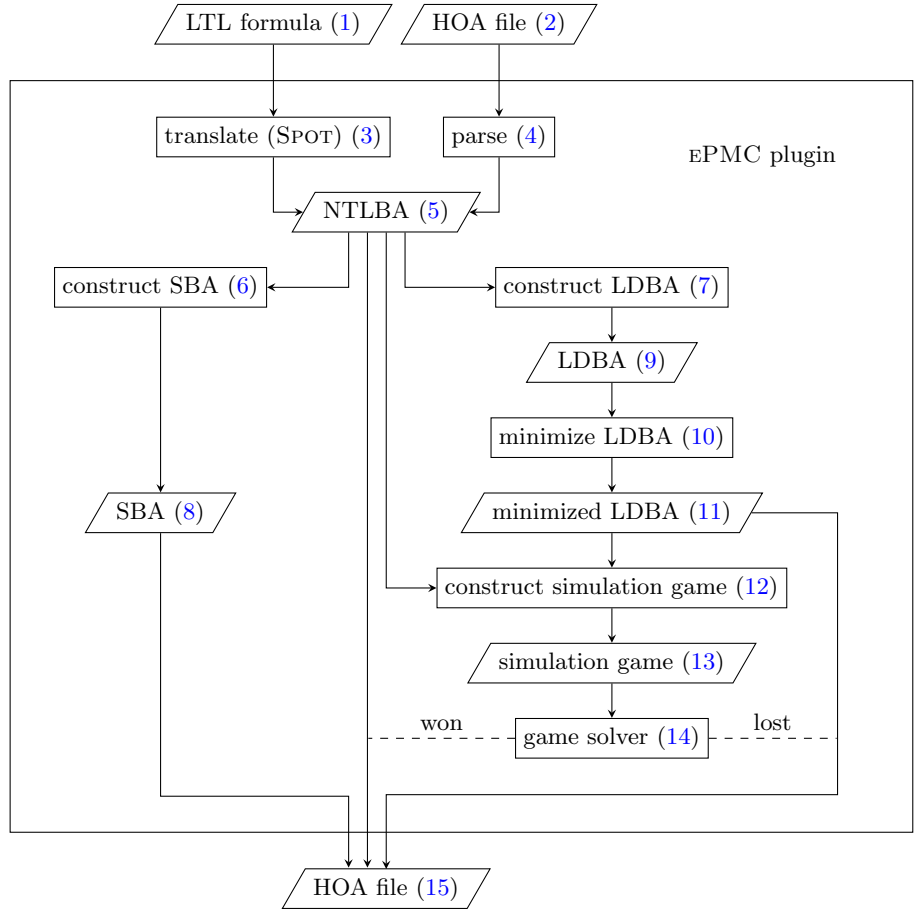

We have implemented slim B\"uchi automata generation as a plugin of the probabilistic model checker \textsc{ePMC} \cite{Hahn14}.
The process is described in Fig.~\ref{fi:epmc}.
The starting point is a transition-labelled B\"uchi automaton in HOA format~\cite{Babiak15} \refnode{hoai} or an LTL formula \refnode{ltl}.
In case we are given an automaton in HOA format, we parse this automaton \refnode{parsehoa} and if we are given an LTL formula, we use the tool \textsc{Spot}~\cite{duret.16.atva2} to transform the formula into an automaton \refnode{ltl2ntba}.
In both cases, we end up with a transition-labelled B\"uchi automaton~\refnode{genba}.

Afterwards, we have two possibilities.
The first option is to transform \refnode{genslim} this automaton into a slim B\"uchi automaton \refnode{sba}~\cite{Hahn20}.
These automata can then be directly composed with MDPs for model
checking or used to produce rewards for learning.

The other option is to construct \refnode{genld} a suitable limit-deterministic B\"uchi automaton (SLDBA)~\refnode{ldba}.
Automata of this type consist of an initial part and a final part.
A nondeterministic choice only occurs when moving from the initial to the final part by an $\varepsilon$ transition (a transition without reading a character).
SLDBA can be directly composed with MDPs.
However, SLDBA directly constructed from general B\"uchi automata are often quite large, which in turn also means that the product with MDPs would be quite large as well.
Therefore, we have implemented further optimization steps.
We can apply a number of algorithms to minimize \refnode{minld} this automaton so as to achieve a smaller SLDBA \refnode{mindld}.
To do so, we implemented several methods:
\begin{itemize}
\item Subsuming the states in the final part with an empty language
\item Signature-based strong bisimulation minimization in the final part
\item Signature-based strong bisimulation minimization in the initial part
\item Language-equivalence of states in the final part
\item If we have a state $s$ in the initial part for which we find a state $s'$ in the final part where the language of $s$ and $s'$ are the same, we can remove all transitions of $s$ and add an $\varepsilon$ transition from $s$ to $s'$ instead.
  Automaton states which cannot be reached anymore afterwards can be removed.
\end{itemize}
Each of these methods has a different potential for minimization as well as runtime:
We allow to specify which optimizations are to be used and in which order they are applied.

Once we have optimized the SLDBA, we could directly use it for later composition with an MDP.
Another possibility is to prove that the original automaton is already suitable for MDPs.
If this is the case, then it is often preferable to use the original automaton:
being constructed by specialized tools such as \textsc{Spot}, it is often smaller than the minimized SLDBA.

The original automaton is suitable if it \emph{simulates} the SLDBA \cite{Hahn20}.
If it does, then it is also composable with MDPs.
Otherwise, it is unknown whether it is suitable for MDPs.
In this case, sometimes more complex notions of simulation can be used.

To show simulation, we construct \refnode{consgame} a simulation game, which in our case is a transition-labelled parity game \refnode{game} with 3 colors.
We solve these games using (a slight variation of) the McNaughton algorithm~\cite{McNaughton66}.
(We are aware that specialized algorithms for parity games with 3 colors exist~\cite{Etessa01}. However, so far the construction of the arena, not solving the game, turned out to be the bottleneck here).
If the even player is winning, the simulation holds.
Otherwise, more complex notions of simulation can be used, which however lead to larger parity games being constructed.
In case the even player is winning for any of them, we can use the original automaton, otherwise we have to use the SLDBA.
In any case, we export the result to an HOA file \refnode{hoao}.
For illustration and debugging purposes, automata and simulation games can also be exported to the GraphViz format~\cite{EllsonGKNW04}.

\iffalse
\begin{itemize}
\item epsilon transitions
\item output automata can either use epsilon transitions or not
\end{itemize}

for future work section
\begin{itemize}
\item better integration with main part of \M{}
\item use better data structures for automata
\item optimization for slim automata
\item visualisation
\end{itemize}
\fi

%%% Local Variables:
%%% mode: latex
%%% TeX-master: "main.tex"
%%% TeX-PDF-mode: t
%%% End:

\section{Two Use Cases}
\label{sec:case-studies}

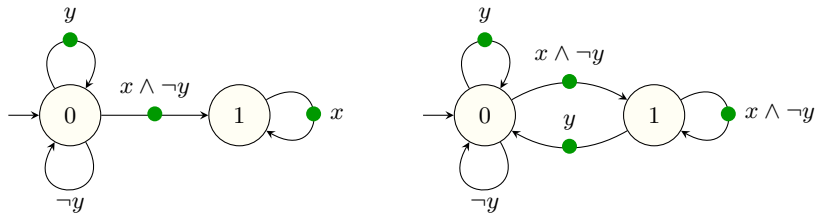
\begin{figure}
  \centering
  \begin{tikzpicture}[every text node part/.style={align=center}]
    % Nodes.
    \node[state,initial,fill=safecellcolor] (Q0) {$0$};
    \node[state,fill=safecellcolor]  (Q1) [right=2.25cm of Q0] {$1$};
    % Transitions.
    \path[->]
    (Q0) edge [loop above] node [accepting dot,label={$y$}] {} ()
    edge[loop below] node {$\neg y$} ()
    edge node[accepting dot,label={$x \wedge \neg y$}] {} (Q1)
    (Q1) edge [loop right] node[accepting dot,label={right: $x$}] {} ();
  \end{tikzpicture}
  \hspace{5mm}
  \begin{tikzpicture}[every text node part/.style={align=center}]
    % Nodes.
    \node[state,initial,fill=safecellcolor] (Q0) {$0$};
    \node[state,fill=safecellcolor]  (Q1) [right=2.25cm of Q0] {$1$};
    % Transitions.
    \path[->]
    (Q0) edge [loop above] node [accepting dot,label={$y$}] {} ()
    edge[loop below] node {$\neg y$} ()
    edge[bend left] node[accepting dot,label={$x \wedge \neg y$}] {} (Q1)
    (Q1) edge [loop right] node[accepting dot,label={right: $x \wedge
      \neg y$}] {} ()
    edge[bend left] node[accepting dot,label={$y$}] {} (Q0);
  \end{tikzpicture}
  \caption{Equivalent, but not equally effective, B\"uchi automata.}
  \label{fi:equivalent-buechi}
\end{figure}

\paragraph{Comparing Automata.}
An $\omega$-regular objective may be described by different automata,
many of which may be good-for-MDPs.  \M{} can be used to compare
the effectiveness of such automata when used in reinforcement
learning.  Consider the two nondeterministic B\"uchi automata shown in
Fig.~\ref{fi:equivalent-buechi}.  Both are equivalent to the LTL
formula $(\eventually\always x) \vee (\always\eventually y)$, but the
one on the right should be better for learning: long transient
sequences of observations that satisfy $x \wedge \neg y$ may convince
the agent to commit to State~1 of the left automaton too soon.

To test this conjecture, we run \M{} on a model with $108$ decision
nodes organized in two long chains.  In one of them the agent sees
many $x$s for a while, but eventually only sees $y$s.  In the other
chain the situation is reversed.  Which chain is followed is up to
chance.  With 2000 episodes, the right automaton
allows $Q$-learning to reliably find an optimal strategy (the
objective is achieved with probability $1$) while the left automaton
most of the time learns strategies that achieve the objective with
probability $0.5$.

\begin{figure}
  \centering
  \begin{tikzpicture}
    \fill[yellow!5] (0,0) rectangle (3,3);
    \draw[step=0.5cm] (0,0) grid (3,3);
    \foreach \x in {0,...,5} {
      \path (0.25+0.5*\x,-0.25) node {\scriptsize$\x$};
      \path (-0.25,0.25+0.5*\x) node {\scriptsize$\x$};
    }
    \fill[blue!40] (1.5,1.5) rectangle (2,2);
    \coordinate (m) at (0.75,0.75);
    \node[red] at (m) {$m$};
    \draw[red,thick,->] (m) +(0.2,0) -- +(0.5,0);
    \draw[red,thick,->] (m) +(0.5,0) -- +(1,0);
    \draw[red,thick,->] (m) +(0,0.2) -- +(0,0.5);
    \draw[red,thick,->] (m) +(0,0.5) -- +(0,1);
    \draw[red,thick,->] (m) +(-0.2,0) -- +(-0.5,0);
    \draw[red,thick,->] (m) +(0,-0.2) -- +(0,-0.5);
    \coordinate (M) at (1.75,2.75);
    \node[darkgreen] at (M) {$M$};
    \draw[darkgreen,thick,->] (M) +(0.2,0) -- +(0.5,0);
    \draw[darkgreen,thick,->] (M) +(0.5,0) -- +(1,0);
    \draw[darkgreen,thick,->] (M) +(-0.2,0) -- +(-0.5,0);
    \draw[darkgreen,thick,->] (M) +(-0.5,0) -- +(-1,0);
    \draw[darkgreen,thick,->] (M) +(0,-0.2) -- +(0,-0.5);
    \draw[darkgreen,thick,->] (M) +(0,-0.5) -- +(0,-1);
  \end{tikzpicture}
  \hspace{8mm}
  \begin{tikzpicture}[every text node part/.style={align=center}]
    % Nodes.
    \node[state,initial,fill=safecellcolor] (Q0) {$0$};
    \node[state,fill=goodcellcolor]  (Q1) [right=2.5cm of Q0] {$1$};
    % Transitions.
    \path[->]
    (Q0) edge [loop above] node {$\neg\mathtt{trapmn} \wedge \mathtt{trapmx},(2)$\\
      $\neg\mathtt{trapmn} \wedge \neg\mathtt{trapmx} \wedge \neg\mathtt{close},(1)$ \\
      $\neg\mathtt{trapmn} \wedge \neg\mathtt{trapmx} \wedge \mathtt{close},(0)$} ()
    edge node {$\mathtt{trapmn},(1)$} (Q1)
    (Q1) edge [loop right] node {$\top,(1)$} ();
  \end{tikzpicture}
  \caption{A grid-world stochastic game arena (left) and a deterministic
    parity automaton for the objective (right).}
  \label{fi:escape-grid}
\end{figure}
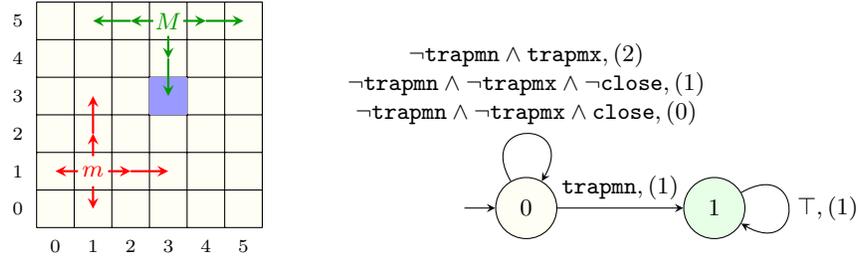

\paragraph{A Game of Pursuit.}
Figure~\ref{fi:escape-grid} describes a stochastic parity game of
pursuit in which the Max player ($M$) tries to escape from the Min
player ($m$). At each round, each player in turn chooses a direction
to move.  If movement in that direction is not obstructed by a wall,
then the player moves either two squares or one square with equal
probabilities.  One square of the grid is a trap, which $m$ must avoid
at all times, but $M$ may visit finitely many times.  Player $M$
should be at least $5$ squares away from player $m$ infinitely often.
This objective is described by the LTL property
$(\eventually \neg\mathtt{trapmn}) \vee ((\eventually \always
\neg\mathtt{trapmx}) \wedge (\always\eventually \neg\mathtt{close}))$,
where $\mathtt{trapmn}$ and $\mathtt{trapmx}$ are true when $m$ and
$M$ visit the trap square, respectively, and $\mathtt{close}$ is true
when the Manhattan distance between the two players is less than $5$
squares.  This objective translates to the deterministic parity
automaton in Fig.~\ref{fi:escape-grid}, which accepts a word if the
maximum recurring priority of its run is odd.

Unlike the example of Fig.~\ref{fi:gambler}, inspection of the Markov
chain induced by an optimal strategy is impractical.  Instead, \M{}
can save the strategy in CSV format.  Postprocessing can then produce
a graphical representation like the one of
Fig.~\ref{fi:escape-strategy}.  The color gradient shows that, in the
main, $M$'s strategy is to move away from $m$.

\input{escape}

\bibliography{papers}
\end{document}